\documentclass[journal]{IEEEtran} 

\usepackage[cmex10]{amsmath}
\usepackage{amssymb}
\usepackage{color}
\usepackage{wasysym} 
\usepackage{booktabs}

\newcommand{\bitem}{\begin{itemize}}
\newcommand{\eitem}{\end{itemize}}

\newcommand{\bpm}{\begin{pmatrix}}   
\newcommand{\epm}{\end{pmatrix}}

\newcommand{\bq}{\begin{equation}}
\newcommand{\eq}{\end{equation}}

\usepackage{graphicx}
\usepackage{bm}
\usepackage{algorithmic}
\usepackage{color}

\usepackage[tight,footnotesize]{subfigure}

\usepackage{epsfig}
\usepackage{epstopdf}

\usepackage[draft]{todonotes}   

\begin{document}
%
 
\title{DeepCut: Object Segmentation from Bounding Box Annotations using Convolutional Neural Networks}

%
\author{Martin Rajchl, Matthew~C.~H.~Lee, Ozan~Oktay, Konstantinos~Kamnitsas, Jonathan~Passerat-Palmbach, Wenjia~Bai, Mellisa Damodaram, Mary A Rutherford, Joseph V Hajnal, Bernhard~Kainz, and~Daniel~Rueckert
\thanks{MR, ML, OO, KK, JPP, WB, BZ and DR are with the Dept. of Computing, Imperial College London, UK. MAR, JVH are with the Dept. of Biomedical Engineering, King's College London, UK. MD is with the Queen Charlotte's Fetal Medicine Department, Hammersmith Hospital and Imperial College London, UK.}
\thanks{(*)Corresponding author: Martin Rajchl e-mail: m.rajchl@imperial.ac.uk}
}


\maketitle
\begin{abstract}
In this paper, we propose \emph{DeepCut}, a method to obtain pixelwise object segmentations given an image dataset labelled weak annotations, in our case bounding boxes. It extends the approach of the well-known \emph{GrabCut} \cite{rother2004grabcut} method to include machine learning by training a neural network classifier from bounding box annotations. We formulate the problem as an energy minimisation problem over a densely-connected conditional random field and iteratively update the training targets to obtain pixelwise object segmentations. Additionally, we propose variants of the \emph{DeepCut} method and compare those to a na{\"i}ve approach to CNN training under weak supervision. We test its applicability to solve brain and lung segmentation problems on a challenging fetal magnetic resonance dataset and obtain encouraging results in terms of accuracy.
\end{abstract}
\begin{IEEEkeywords}
Bounding Box, Weak Annotations, Image Segmentation, Machine Learning, Convolutional Neural Networks, DeepCut
\end{IEEEkeywords}

\IEEEpeerreviewmaketitle

\section{Introduction}
\PARstart{M}{any} modern medical image analysis methods that are based on machine learning rely on large amounts of annotations to properly cover the variability in the data (\emph{e.g.} due to pose, presence of a pathology, \emph{etc}). However, the effort for a single rater to annotate a large training set is often not feasible. To address this problem, recent studies employ forms of weak annotations (\emph{e.g.} image-level tags, bounding boxes or scribbles) to reduce the annotation effort and aim to obtain comparably accurate results as to under full supervision (\emph{i.e.} using pixelwise annotations) \cite{papandreou2015weakly,dai2015boxsup,schlegl2015predicting}. 

User-provided bounding boxes are a simple and popular form of annotation and have been extensively used in the field of computer vision to initialise object segmentation methods \cite{rother2004grabcut,lempitsky2009image}. Bounding boxes have advantages over other forms of annotations (e.g. scribbles or brush strokes \cite{boykov2001interactive, rajchl2014interactive, baxter2015optimization}), as they allow to spatially constrain the problem (\emph{i.e.} ideally, the object is unique to the bounding box region and fully contained in it). In a practical sense, bounding boxes can be defined via two corner coordinates, allowing fast placement (approximately 15 times faster than pixelwise segmentations \cite{lin2014microsoft}) and lightweight storage of the information. Considering the required interaction effort and the amount of provided information, these properties qualify bounding boxes as preferred weak annotation for image analysis methods. 

In segmentation studies \cite{rother2004grabcut,lempitsky2009image,cheng2015densecut}, bounding boxes are employed as both initialisation and spatial constraints for the segmentation problem. The above approaches model the image appearance (\emph{i.e.}, colours or greyscale intensity) and impose smoothness constraints upon the segmentation results for each image. However, given an image database and corresponding annotations, we can assume that objects share common shape and appearance information, which can be learned (\emph{i.e.}, instead of direct image-by-image object segmentation, a common model can be learned for the all images in the database). This is particularly interesting for segmentation problems on medical images, where typically an entire cohort is to be analysed for a specific organ or region, exhibiting large class similarity in terms of shape and appearance.

In this paper, we propose combining a neural network model with an iterative graphical optimisation approach to recover pixelwise object segmentations from an image database with corresponding bounding box annotations. The idea builds on top of the popular GrabCut \cite{rother2004grabcut} method, where an intensity appearance model is iteratively fitted to a region and subsequently regularised to obtain a segmentation. Similarly to this, the proposed \emph{DeepCut} method iteratively updates the training targets (\emph{i.e.} the class associated with a voxel location, described by an image patch) learned by a convolutional neural network (CNN) model and employs a fully connected conditional random field (CRF) to regularise the segmentation. The approach is formulated in a generic form and thus can be readily applied to any object or image modality. We briefly review recent advancements in the following section to put this approach into the context of the current state-of-the-art and highlight our contributions.

\subsection{Related work}
Graphical energy minimisation techniques are popular methods for regularised image segmentation due to inherent optimality guarantees and computational efficiency \cite{boykov2001fast,komodakis2007fast}. They have been extensively used in the optimisation of interactive \cite{boykov2001interactive,freedman2005interactive,price2010geodesic, baxter2015optimization} and fully automated segmentation methods \cite{xia2013semantic,wolz2010measurement,koch2015multi,rajchl2016hierarchical}.

An iterative optimisation of such energy functionals allows to address more complex problems, such as the pixelwise segmentation from bounding box annotations. The popular GrabCut method \cite{rother2004grabcut} iteratively updates parameters of a Gaussian mixture model (GMM) and optimises the resulting energy with a graph cut. Lempitsky \emph{et al.}  \cite{lempitsky2009image} extended this method by adding a topological prior to prevent excessive shrinking of the region and Cheng et al. \cite{cheng2015densecut} improved upon its performance by employing a fast fully connected Conditional-Random Field (CRF) \cite{krahenbuhl2012efficient}. Similar approaches include the time-implicit propagation of levelsets via continuous max-flow \cite{yuan2013jointly,rajchl2015variational}, iterative graph cuts \cite{nambakhsh2013left} and the use of the expectation-maximisation (EM) algorithm \cite{dempster1977maximum,kapur1996segmentation,papandreou2015weakly}.

Similarly to the above mentioned segmentation approaches, learning-based methods have been recently investigated to exploit the advantages of weak annotations, primarily to reduce the effort of establishing training data. In contrast to learning under full supervision (\emph{i.e.} using pixelwise annotations), weakly supervised methods aim to learn from image-level tags, partial labels, bounding boxes, \emph{etc.} and infer pixelwise segmentations.

Recently, several multiple instance learning (MIL) techniques were investigated, particularly when images could potentially contain multiple objects. Cinbis \emph{et al.} \cite{cinbis2014multi} proposed a multi-fold MIL method to obtain segmentations from image level tags. Vezhnevets and Buhmann \cite{vezhnevets2010towards} addressed the problem with a Texton Forest \cite{shotton2008semantic} framework, extending it to MIL. With the re-emerging of convolutional neural networks \cite{ciresan2012multi,krizhevsky2012imagenet}, MIL methods have been proposed to exploit such methods \cite{pathak2014fully,pinheiro2015image}. However MIL-based methods, even when including additional modules under weak supervision \cite{pinheiro2015image}, have not been able to achieve comparable accuracy to fully supervised methods \cite{papandreou2015weakly}. 
However, latest developments using CNN learning with weakly supervised data have shown remarkable improvements in accuracy. Schlegl et al. \cite{schlegl2015predicting} parse clinical reports to associate findings and their locations with optical coherence tomography images and further obtain segmentations of the reported pathology. Dai et al. \cite{dai2015boxsup} iterative between updating region proposals in bounding boxes and model training. Papandreou et al. \cite{papandreou2015weakly} formulate an Expectation-Maximization (EM) algorithm \cite{dempster1977maximum} to iteratively update the training targets. Both of the latter methods were able to achieve comparable accuracy to those under full supervision. 

\subsection{Contributions}
In this paper, we build upon the ideas of GrabCut \cite{rother2004grabcut}, a well-known object segmentation method employed on single images. We extend the basic idea with recent advances in CNN modelling and propose \emph{DeepCut}, a method to recover semantic segmentations given a database of images with corresponding bounding boxes. For this purpose, we formulate an iterative energy minimisation problem defined over a densely connected conditional random field (CRF) \cite{krahenbuhl2012efficient} and use it to update the parameters of a CNN model. We compare the proposed method against a fully supervised ($\text{CNN}_\text{FS}$) and a na{\"i}ve approach to weakly supervised segmentation ($\text{CNN}_\text{na{\"i}ve}$), to obtain upper and lower accuracy bounds for a given segmentation problem. Further, we examine the effect of region initialisation on the proposed method by providing a pre-segmentation within the bounding box, ($\text{DC}_\text{PS}$)). Finally, we compare all methods in their segmentation accuracy using a highly heterogeneous and challenging dataset of fetal magnetic resonance images (MRI) and further evaluate the performance to GrabCut \cite{rother2004grabcut}, as an external method to this framework.

\section{Methods}
Let us consider segmentation problems using energy functionals over graphs as described in \cite{boykov2001fast}. We seek a labelling $f$ for each pixel $i$, minimising 

\begin{equation}
\begin{aligned}
E(f) = \sum_{i} \psi_{u} (f_i) + \sum_{i<j} \psi_{p} (f_i, f_j) \, ,\\
\end{aligned}
\label{eq:gen_energy}
\end{equation}

where $\psi_{u} (f_i)$ serves as unary data consistency term, measuring the fit of the label $f$ at each pixel $i$, given the data. Additionally, the pairwise regularisation term $\psi_{p} (f_i, f_j)$ penalises label differences for any two pixel locations $i$ and $j$. Typically, pairwise regularisation terms have the form of 

\begin{equation}
\begin{aligned}
\psi_{p} (f_i, f_j) \propto \exp \left( - \frac{(I_i - I_j)^2}{2\theta_\beta} \right)\, ,\\
\end{aligned}
\label{eq:gen_pw}
\end{equation}

and enforce contrast-sensitive smoothness penalties between the intensity vectors $I_i$ and $I_j$ \cite{boykov2001interactive,koch2015multi}. We can minimise the energy in Eq. \eqref{eq:gen_energy} using a densely-connected CRF \cite{krahenbuhl2012efficient}, where we replace the pairwise term with

\begin{equation}
\begin{aligned}
\psi_{p}(f_i) = g(i,j) [f_i \neq f_j] \, ,
\end{aligned}
\end{equation}

consisting of two penalty terms modelling appearance \eqref{eq:appearance_kernel} and smoothness \eqref{eq:smoothness_kernel} between the locations $p_i$ and $p_j$:

\begin{subequations}
\begin{align}
g(i,j) & = \omega_1 \exp \left( - \frac{|p_i - p_j|^2}{2 \theta_\alpha^2}  - \frac{|I_i - I_j|^2}{2 \theta_\beta^2} \right) \label{eq:appearance_kernel} \\
& + \omega_2 \exp \left( - \frac{|p_i - p_j|^2}{2 \theta_\gamma^2} \right)  \, . \label{eq:smoothness_kernel} 
\end{align}
\end{subequations}

The relative contributions of the two penalties is weighted by the regularisation parameters $\omega_1$ and $\omega_2$ and the degrees of spatial proximity and similarity are controlled by $\theta_\alpha$ and $\theta_\beta$, respectively \cite{krahenbuhl2012efficient}. \\

The unary potential is computed independently for each pixel $i$ by a data model with the parameters $\Theta$ that produces a distribution $y_i$ over the label assignment given an input image or patch $x$ and is defined as the negative log-likelihood of this probability:

\begin{equation}
\begin{aligned}
\psi_{u}(f_i) = - \log P(y_i | x, \Theta) \, . 
\end{aligned}
\label{eq:data_term}
\end{equation}

In contrast to \cite{rother2004grabcut}, where the unary term is computed from a GMM of the observed colour or intensity vector, we employ a CNN with the parameters $\Theta$. We describe the network architecture in Section \ref{sec:cnn} in detail.

\subsection{Segmentation by Iterative Energy Optimisation} \label{sec:iterative}
The proposed \emph{DeepCut} method can be seen as an iterative energy minimisation method similar to GrabCut \cite{rother2004grabcut}. There are two key stages to both algorithms, model estimation and label update. GrabCut uses a GMM to parametrise the colour distributions of the foreground and background. In the model estimation stage the parameters $\Theta$ for the GMM are computed based on the current label assignment $f$ for each pixel $i$. At the label update stage the pixels are relabelled based on the new model. \emph{DeepCut} replaces the GMM with a Neural Network model and the graph cut solver from \cite{boykov2001fast} with \cite{krahenbuhl2012efficient} on a densely-connected graph. In contrast to \cite{rother2004grabcut}, and rather than recomputing our model, we make use of transfer learning \cite{oquab2014learning} and reinitialise the CNN with the parameters of the last iteration.

This two-step iterative method is similar to an EM algorithm, consisting of an E-step (label update) and M-step (model update). Papandreou et al. \cite{papandreou2015weakly} describe such a method to iteratively update $f$, however only employ regularisation as in Eq. \eqref{eq:gen_energy} as a post-processing step during testing.

\subsection{Convolutional Neural Network Model} \label{sec:cnn}
\begin{figure}
\centering
\includegraphics[width=0.75\linewidth]{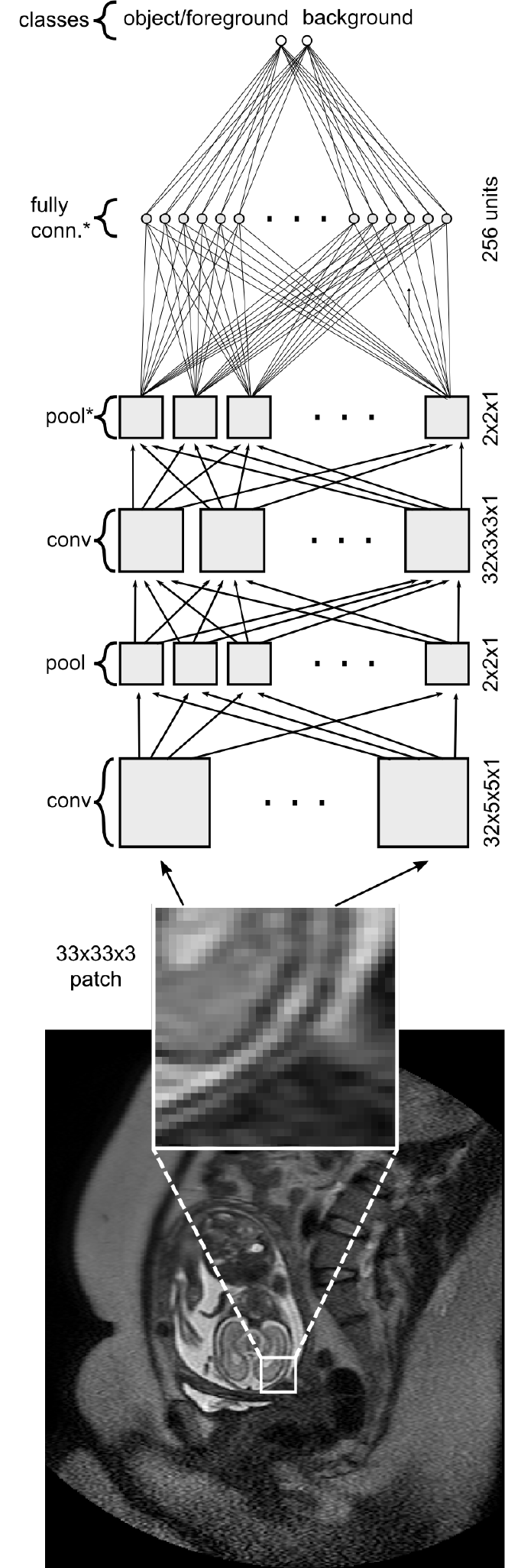}
\vspace{-2mm}
\caption{\label{fig:net_arch} CNN architecture with convolutional (conv), max-pooling (pool) layers, and fully connected layers for foreground/background classification. Layer, which inputs are subjected to 50\% dropout \cite{srivastava2014dropout} are marked with *.}
\end{figure}

The CNN is a hierarchically structured feed-forward neural network comprising of least one convolutional layer \cite{ciresan2012multi,krizhevsky2012imagenet}. Additionally, max-pooling layers downsample the input information to learn object representations at different scales. Downstream of several convolutional and max-pool layers is typically a layer of densely-connected neurons, reducing the output to a desired number of classes \cite{schlegl2015predicting}. Our CNN is a typical feed-forward neural network model which consist of convolutional layers (feature extraction), max-pooling layers (shift/scale invariance) and dense layers (classification stage). The network architecture used in this study is depicted in Fig. \ref{fig:net_arch}. \\

\subsubsection*{Input \& Output Space}
Given a database of size $N$ containing images $I = \{I_1,...,I_N\}$ and corresponding bounding boxes $B = \{B_1,...,B_N\}$, we attempt to learn pixelwise segmentations of the objects depicted in $I$ and constrained to $B$. For this purpose, we employ a CNN with parameters $\Theta$ to classify image patches centred around a voxel location $i$ into foreground and background. 

We describe each voxel location $i \in I_n$ as a 3D patch of size $p_x \times p_y \times p_z$, centred around $X$. Each patch $X$ is associated with an integer value $Y = \{0,1\}$, representing background and foreground class of the centre voxel at $i$, respectively. The patch $X$ serves as input to the network, which aims to predict $Y$. Because of anticipated motion artefacts between fetal MR slices, we decided to emphasise the in-plane context of our training patches, using a patch size of 33x33x3, rather than cubic patch dimensions. \\

\subsubsection*{Network Configuration} \label{sec:cnn_config}
For the purpose of this study, we designed a simple CNN inspired by the well-known \emph{LeNet} architecture \cite{lecun1989backpropagation}. While the proposed approach can employ other networks, we chose this configuration, as it is easily understandable and simple to reproduce. The CNN consists of two sets of convolutional (\emph{conv}) and subsequent max-pooling (\emph{pool}) layers. Since a convolution reduces the layer output by half the kernel size, we pad its output with zeros to preserve the size of the input tensor. The two serial \emph{conv/pool} layers are attached to a layer with densely-connected (\emph{dense}) neurons and an output layer with neurons associated with foreground and background. For regularisation purposes, we add dropout to both inputs of the \emph{dense} and output layer, randomly sparsifying the signal and reducing the potential of over-fitting \cite{srivastava2014dropout}. Figure \ref{fig:net_arch} depicts the specific configuration, which has been fixed for all experiments in this study.\\

\subsubsection*{Training \& Optimisation}
In each epoch, we extract $K = 10^5$ patches $X_k$ from the training database, which are equally distributed between the classes $Y_k$. All CNN weights are initialised by sampling from a Gaussian distribution. We employ an adaptive gradient descent optimisation (ADAGRAD) \cite{duchi2011adaptive} of mini-batches using a constant learning rate $\eta = 0.015$ for a fixed number of epochs. This has the advantage of adapting the learning rate locally for each feature, thus exhibiting more robust behaviour and faster convergence. The loss function is defined as the categorical cross-entropy between the true and coding distributions, respectively.\\

\subsubsection*{Data Augmentation}
The training set undergoes data augmentation for better generalisation of the learned features and to prevent over-fitting to the training data. For this purpose, we add a Gaussian-distributed intensity offset to each patch with the standard deviation $\sigma$ and randomly flip the patch across the spatial dimensions to increase the variation in the training data.

\subsection{Na{\"i}ve Learning Approach} \label{sec:naive}
\begin{figure*}
\centering
\includegraphics[width=0.95\linewidth]{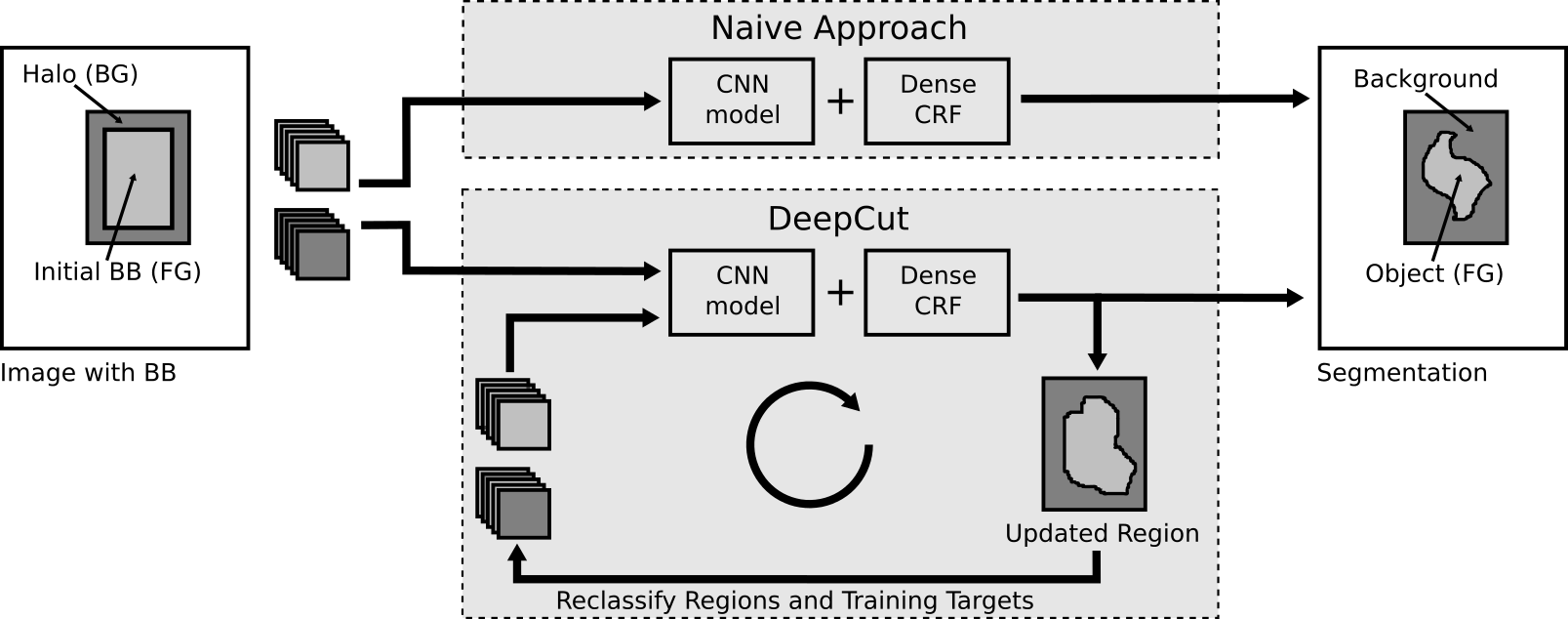}
\vspace{-2mm}
\caption{\label{fig:deepCut} Na{\"i}ve CNN learning versus the proposed \emph{DeepCut} approach, iteratively updating the learning target classes for input patches.}
\end{figure*}

If we assume that the patches describing the object constrained to the bounding box $B$ are unique, we can attempt to classify patches $X_i \in B$ into object and background, by using patches sampled from the bounding boxes as foreground targets. To obtain background targets, we can extend $B$ to a halo region $H$, solely containing background voxels. 
A na{\"i}ve approach to segmentation would be to assume that all $X_i \in B$ belong to the object and all $X_i \in H$ to the background. However, since the region $B$ contains false positive locations (\emph{i.e.}, the object does not fully extend to the entire bounding box region, but is merely a subset of it), we will introduce errors into the model, which will impact the accuracy of the final segmentation. To ensure a fair comparison of the segmentation results, post-processing includes regularisation with a densely-connected CRF \cite{krahenbuhl2012efficient}. This na{\"i}ve approach ($\text{CNN}_\text{na{\"i}ve}$) is depicted in Figure \ref{fig:deepCut}.

\subsection{DeepCut}
In order to develop a better approach compared to Section \ref{sec:naive}, we start training the CNN model $\Theta$ with patches $X$ sampled from $B$ and $H$ for foreground and background, respectively. In contrast to the na{\"i}ve approach, we interrupt the training of $\Theta$ after a fixed number of epochs and update the classes $Y$ for all voxels in $B = R_{FG} \cup R_{BG}$, via inference and subsequent CRF regularisation, according to Section \ref{sec:iterative}. We continue training with the updated targets and reinitialise the CNN with $\Theta$.

\subsection{Region Initialisation} \label{sec:init}
While methods such as \emph{DeepCut}, GrabCut \cite{rother2004grabcut} or others \cite{papandreou2015weakly,dai2015boxsup} rely on (approximately) globally optimal solvers (\emph{i.e.}, \cite{boykov2001fast} and \cite{krahenbuhl2012efficient}, respectively), the iterative nature of the algorithm will be limited to finding local optima. Thus the resulting segmentation is dependent on the initial regions $R_{FG}$ and $R_{BG}$. Papandreou et al. \cite{papandreou2015weakly} propose performing a pre-segmentation within $B$ to initialise $R_{FG}$ and $R_{BG}$ closer to the object and observed large improvements in accuracy of the final segmentation. Similar initialisation can be done with the proposed \emph{DeepCut} method and will be examined in Section \ref{sec:experiments}. In our experiments, we distinguish the variants of \emph{DeepCut} initialised with bounding boxes and pre-segmentations with $\text{DC}_\text{BB}$ and $\text{DC}_\text{PS}$, respectively.

\section{Experiments} \label{sec:experiments}

\subsection{Image Data}
For all experiments, we used the database in \cite{damodaram2012foetal}, consisting of MR images of 55 fetal subjects. The images were acquired on 1.5T MR scanner using a T2-weighted ssFSE sequence (scanning parameters: TR 1000ms, TE 98ms, 4mm slice thickness, 0.4mm slice gap). Most of the images contain motion artefacts that are typical for such acquisitions. The imaged population consists of 30 healthy subjects and 25 subjects with intrauterine growth restriction (IUGR) and their gestational age ranged from 20 to 30 weeks. For all images, the brain and the lung regions have been manually segmented by an expert rater. We want to emphasise that the brain segmentations are in fact not tissue segmentations, but whole brains similar to the data used in \cite{keraudren2014automated,taleb2013automatic}.

\subsection{Preprocessing \& Generation of Bounding Boxes}
All images underwent bias field correction \cite{sled1998nonparametric} and normalisation of the image intensities to zero mean and unit standard deviation computed from the bounding box. Bounding boxes $B$ were generated from manual segmentations by computing the maximum extent of the segmentation and enlarging it by 5 voxels for each slice. The background halo regions $H$ were created by extending $B$ by 20 voxels. 

\subsection{Comparative Methods}
To compare the performance of the proposed \emph{DeepCut} approach, we fix the CNN architecture (see  Section \ref{sec:cnn_config}), preprocessing and CRF parameters for all learning-based methods. 
We can consider the $\text{CNN}_\text{na{\"i}ve}$ (\emph{c.f.} Section \ref{sec:naive}) a lower bound in terms of accuracy performance. Alternatively, we train the CNN directly under full supervision (\emph{i.e.} from pixelwise segmentations, $\text{CNN}_\text{FS}$) and predict into $B$, which can be considered an upper accuracy bound given model complexity and data. 
We assess both the performance of the proposed \emph{DeepCut} initialised by bounding boxes ($\text{DC}_\text{BB}$) or via a GrabCut pre-segmentation ($\text{DC}_\text{PS}$) as suggested in Section \ref{sec:init}. Lastly, we state results from the GrabCut (GC) method \cite{rother2004grabcut} for a performance comparison external to the proposed framework.

\subsection{Experimental Setup, Evaluation \& Parameter Selection}
We performed 5-fold cross-validation of randomly selected healthy and IGUR subjects and fixed the training and testing databases for all compared methods. The resulting segmentations are evaluated in their overlap with expert manual segmentations using the Dice Similarity Coefficient, measuring the mean overlap between two regions $A$ and $B$: 

\begin{equation}
\begin{aligned}
DSC = \frac{2 | A \cap B | }{|A| + |B|}
\end{aligned}
\label{eq:penalty_1}
\end{equation}

Three datasets were left out of the evaluation experiments and used to tune the GrabCut MRF regularisation weight and the CRF regularisation parameters in \eqref{eq:smoothness_kernel} and \eqref{eq:appearance_kernel} via random permutations of the parameter combinations using manual segmentations.

\begin{table}[h!]\caption{\label{ta:parameters} DeepCut parameters}
\begin{tabular}{ll}
\hline 
\hline
\noalign{\vskip 1mm} 
   Parameter & Value\\
   \noalign{\vskip 1mm} 
\hline
\noalign{\vskip 1mm} 
	\multicolumn{2}{c}{Convolutional Neural Network}  \\
	\noalign{\vskip 1mm} 
\hline
\noalign{\vskip 1mm} 
	Patch size ($p_x$ x $p_y$ x $p_z$) & $33$ x $33$ x $3$ \\ 
    Learning rate $\eta$  & $0.015$ \\ 
    $N_{Epochs}$ (Brain) & $500$ \\ 
    $N_{Epochs}$ (Lungs) & $250$ \\ 
    $N_{Epochs}$ per DeepCut iteration & $50$ \\ 
    $N_{Batch}$ & $10^5$ patches per Epoch \\ 
    $N_{Mini-batch}$ & $5000$ patches \\
    \noalign{\vskip 1mm}  
\hline
\noalign{\vskip 1mm} 
	\multicolumn{2}{c}{Densely-connected CRF}  \\
	\noalign{\vskip 1mm} 
\hline
\noalign{\vskip 1mm} 
	$\omega_1$, $\omega_2$  & 5.0 \\
	$\theta_\alpha$  & 10.0 \\
	$\theta_\beta$ (Brain) & 20.0 \\
	$\theta_\beta$ (Lungs) & 0.1 \\
	$\theta_\gamma$ (Brain) & 1.0 \\
	$\theta_\gamma$ (Lungs) & 0.1 \\
	$N_{Iterations}$ & 5 \\
	\noalign{\vskip 1mm} 
\hline
\noalign{\vskip 1mm} 
\multicolumn{2}{c}{GrabCut (see \cite{rother2004grabcut}) }  \\
\noalign{\vskip 1mm} 
\hline
\noalign{\vskip 1mm} 
	$\gamma$ (Brain) & 2.5 \\
	$\gamma$ (Lungs) & 1.0 \\
	\noalign{\vskip 1mm} 
\hline \hline
\end{tabular}
\end{table}

\subsection{Implementation Details \& Hardware}
We implemented the CNN as shown in Fig. \ref{fig:net_arch} with \emph{Lasagne}\footnote{http://lasagne.readthedocs.org/} and \emph{Theano}\footnote{http://deeplearning.net/software/theano/} \cite{bastien2012theano}. All experiments were run on Ubuntu 14.04 machines with 256 GB memory and a single Tesla K80 (NVIDIA Corp., Santa Clara, CA) with 12GB of memory. We used the CRF implementation in \cite{kamnitsas2015multi} to solve Eq. \eqref{eq:gen_energy}, according to \cite{krahenbuhl2012efficient}.

\section{Results}
\begin{figure*}
\centering
\includegraphics[width=0.95\linewidth]{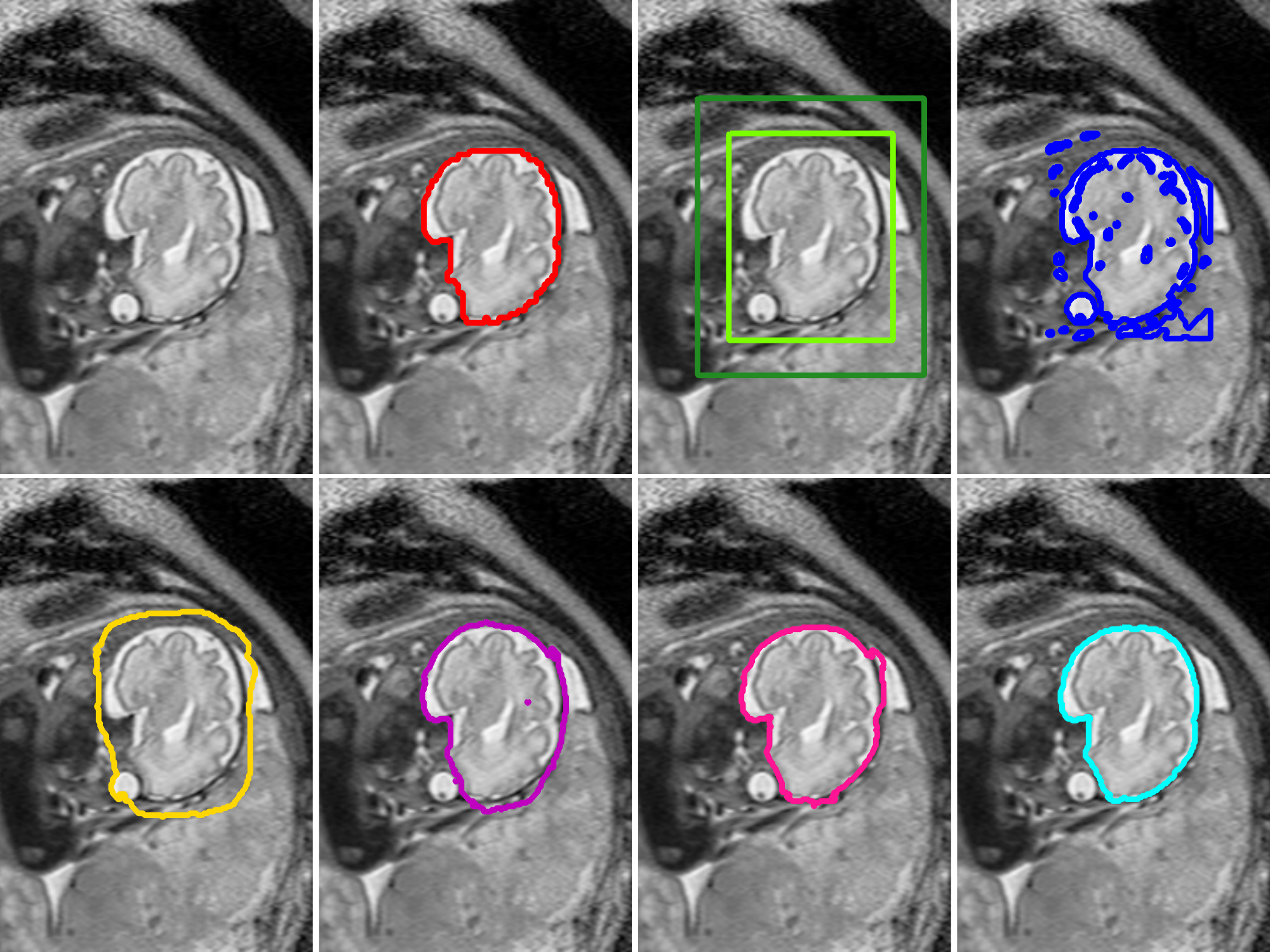}
\vspace{-2mm}
\caption{\label{fig:example_seg_brain} Example brain segmentation results for all compared methods: Top row (from left to right): (1) original image (2) manual segmentation (red), (3) initial bounding box $B$ with halo $H$, (4) GrabCut \cite{rother2004grabcut} (GC, blue). Bottom Row: (5) na{\"i}ve learning approach ($\text{CNN}_\text{na{\"i}ve}$, yellow), (6) DeepCut from bounding boxes ($\text{DC}_\text{BB}$, purple), (7) DeepCut from pre-segmentation ($\text{DC}_\text{PS}$, pink) and (8) fully supervised CNN segmentation ($\text{CNN}_\text{FS}$, cyan).}
\end{figure*}

\begin{figure*}
\centering
\includegraphics[width=0.95\linewidth]{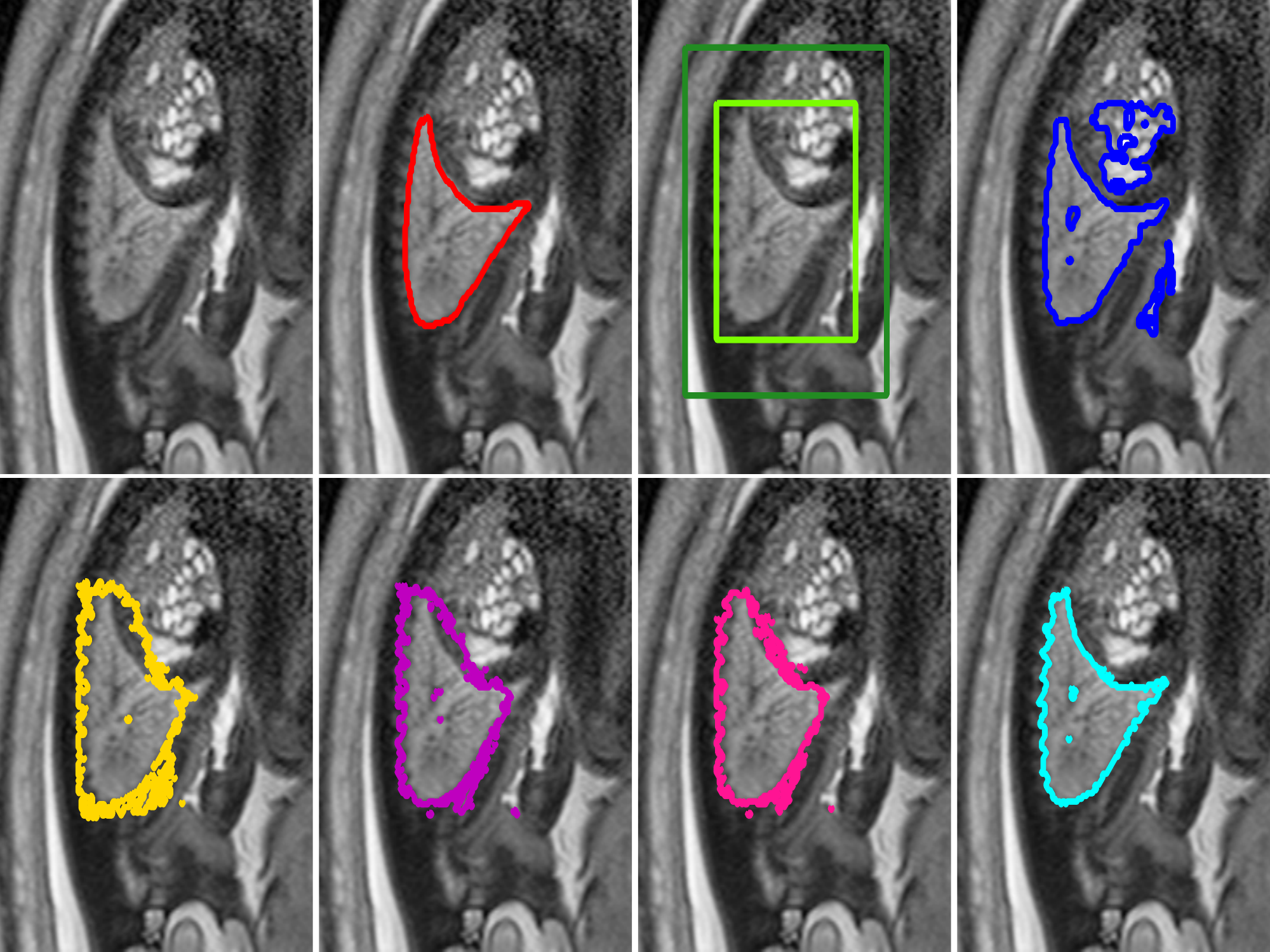}
\vspace{-2mm}
\caption{\label{fig:example_seg_lungs} Example lung segmentation results for all compared methods: Top row (from left to right): (1) original image (2) manual segmentation (red), (3) initial bounding box $B$ with halo $H$, (4) GrabCut \cite{rother2004grabcut} (GC, blue). Bottom Row: (5) na{\"i}ve learning approach ($\text{CNN}_\text{na{\"i}ve}$, yellow), (6) DeepCut from bounding boxes ($\text{DC}_\text{BB}$, purple), (7) DeepCut from pre-segmentation ($\text{DC}_\text{PS}$, pink) and (8) fully supervised CNN segmentation ($\text{CNN}_\text{FS}$, cyan).}
\end{figure*}

Example segmentation results of all compared methods and initialisations can be found in Figures \ref{fig:example_seg_brain} and \ref{fig:example_seg_lungs}. We observe comparable agreement of the proposed $\text{DC}_\text{PS}$ and a fully supervised $\text{CNN}_\text{FS}$ for the brain region. Generally, an increase of segmentation accuracy can be seen in brain and lungs with increasing sophistication of the learning-based methods. 

\subsection{Na{\"i}ve Learning Approach versus \emph{DeepCut}}
Comparison of methods directly learning from bounding boxes (i.e. $\text{CNN}_\text{na{\"i}ve}$ and $\text{DC}_\text{BB}$, demonstrate that the iterative target update of the proposed DeepCut method results in large improvements in accuracy for both the brain and the lungs (see Fig. \ref{fig:example_seg_brain} and \ref{fig:example_seg_lungs}), Tab. \ref{ta:acc_brain} and \ref{ta:acc_lungs}). Numerically, we obtain an increase of 12.6\% and 8.9\% in terms of average DSC for the brain and lungs, respectively.

\subsection{Initialisation with Pre-segmentations}
Further, when a pre-segmentation instead of bounding boxes is used to initialise the \emph{DeepCut} method, the mean DSC is improved by another 3.7\% for the brain and 4.9\% for the lungs. This can be seen in the example segmentation in Fig. \ref{fig:example_seg_brain} and \ref{fig:example_seg_lungs}, where the $\text{DC}_\text{PS}$ segmentation for both organs is visually closer to those of the fully supervised $\text{CNN}_\text{FS}$.

\subsection{Comparison with GrabCut}
While the comparative GrabCut method performs well for the brain (DSC $80.7 \pm 4.9\%$), we observe less robust behaviour in the lungs (DSC $58.6 \pm 19.0\%$). GrabCut outperforms the $\text{CNN}_\text{na{\"i}ve}$ for brain segmentation, however the presence of large outliers results in a lower mean accuracy in the lung regions. Several segmentations present with DSC $<20\%$, indicating that the GrabCut was not able to detect an object in some cases or the segmenting false positive voxels in others.

\subsection{\emph{DeepCut} versus Fully Supervised Learning}
\begin{figure}
\centering
\includegraphics[width=0.95\linewidth]{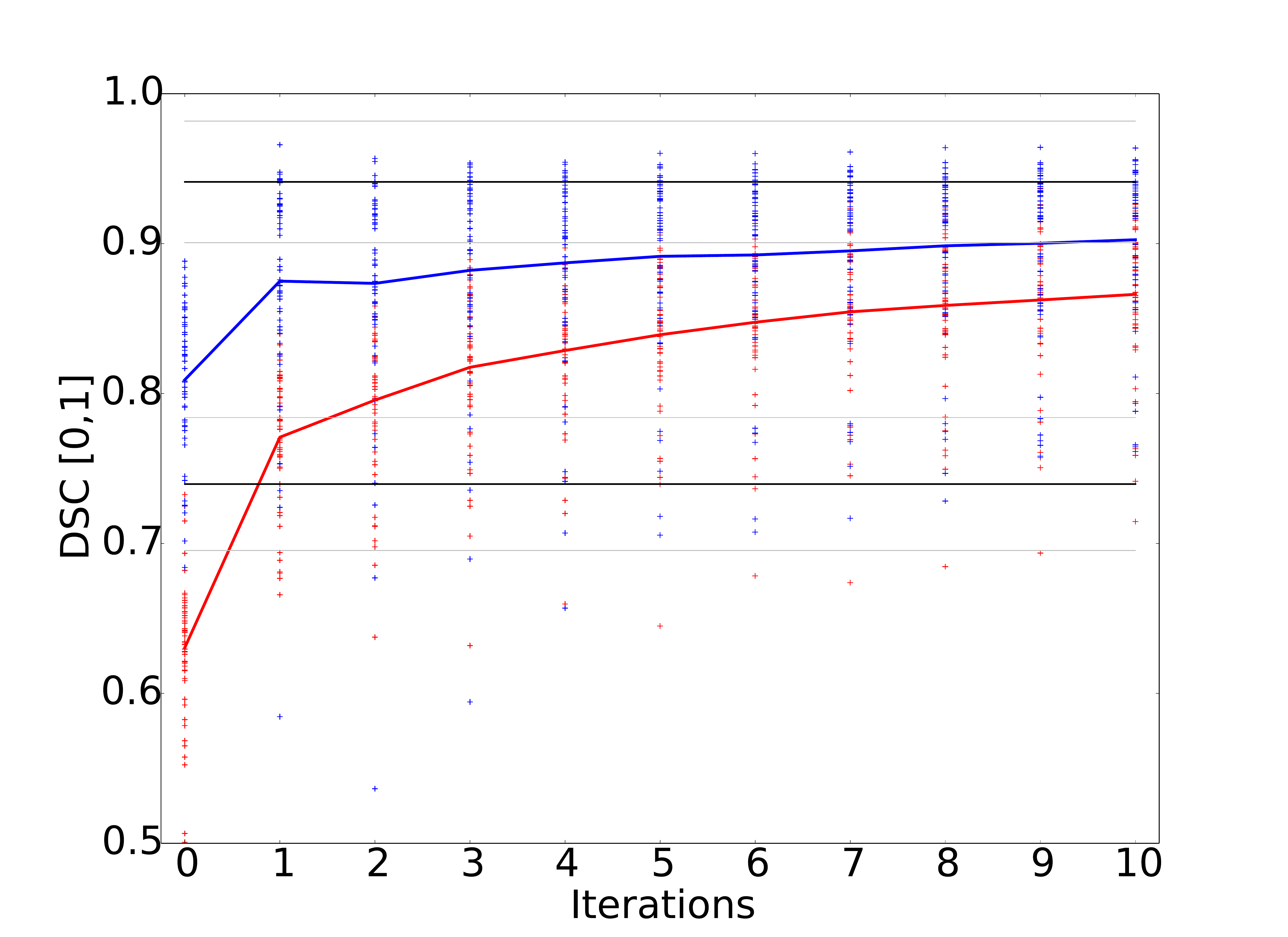}
\vspace{-2mm}
\caption{\label{fig:acc_iterations} Accuracy improvement in terms of DSC over \emph{DeepCut} iterations in case of fetal brain segmentations. DeepCut initialisation with bounding boxes ($\text{DC}_\text{BB}$) (red) versus initialisation with pre-segmentation ($\text{DC}_\text{PS}$) (blue) in context with lower ($\text{CNN}_\text{na{\"i}ve}$) and upper ($\text{CNN}_\text{FS}$) accuracy bound, depicted with mean (black) and standard deviations (grey). }
\end{figure}

We halted training and evaluated intermediate accuracy results of the proposed \emph{DeepCut} variants over iterations. For both $\text{DC}_\text{BB}$ and $\text{DC}_\text{PS}$, accuracy increases after each iteration (see Fig. \ref{fig:acc_iterations}) and approaches the upper bound of fully supervised training ($\text{CNN}_\text{FS}$). Most importantly, both proposed \emph{DeepCut} methods present with higher average accuracy than the na{\"i}ve approach, however the accuracy remains lower than $\text{CNN}_\text{FS}$. For reference, in Fig. \ref{fig:acc_iterations} the mean (black) and standard deviation (gray) of $\text{CNN}_\text{na{\"i}ve}$ and $\text{CNN}_\text{FS}$ are also shown.

\begin{figure*}
    \centering
    \subfigure[Brain]{{\includegraphics[width=0.47\linewidth]{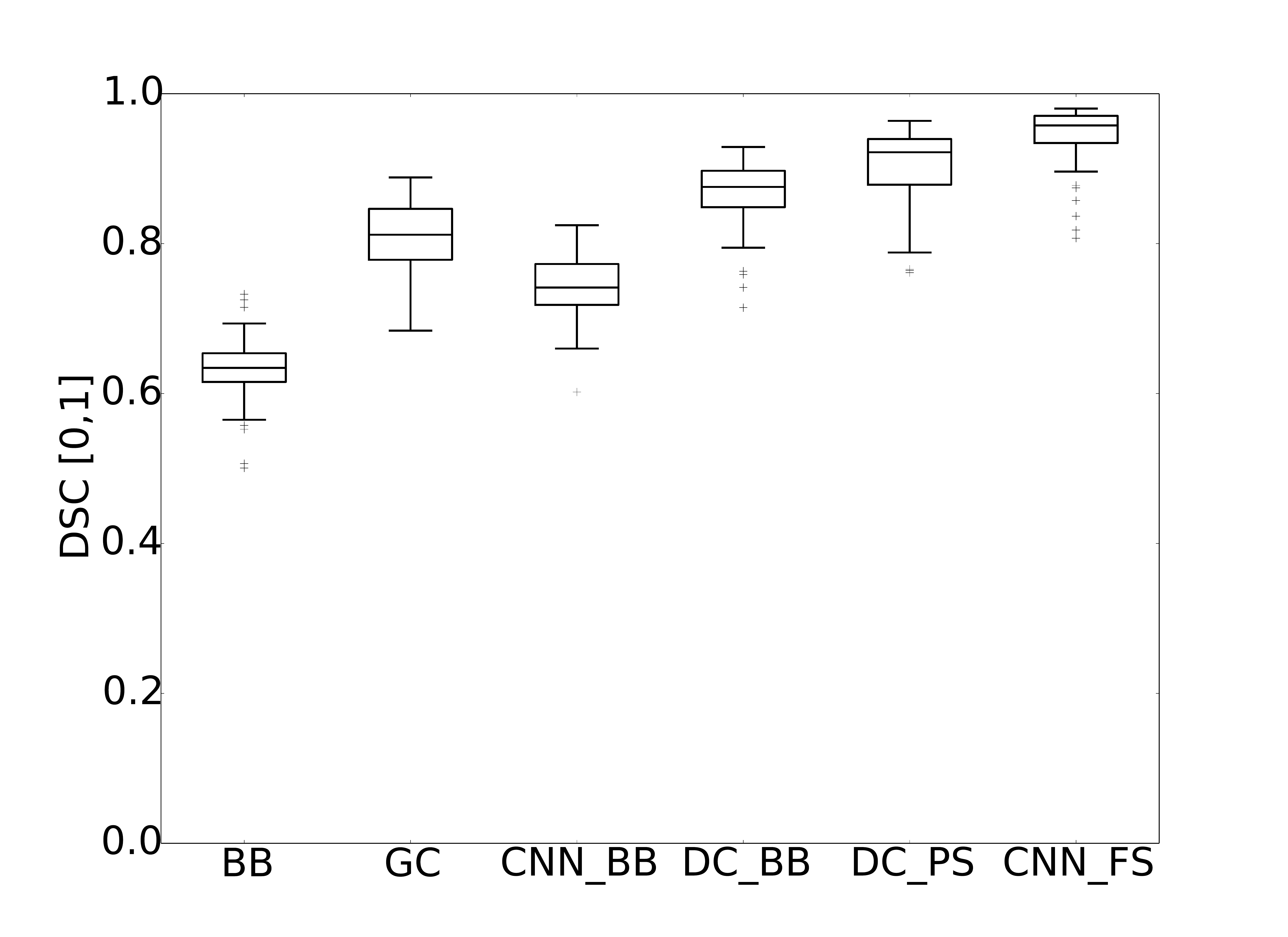} }}%
    \qquad
    \subfigure[Lungs]{{\includegraphics[width=0.47\linewidth]{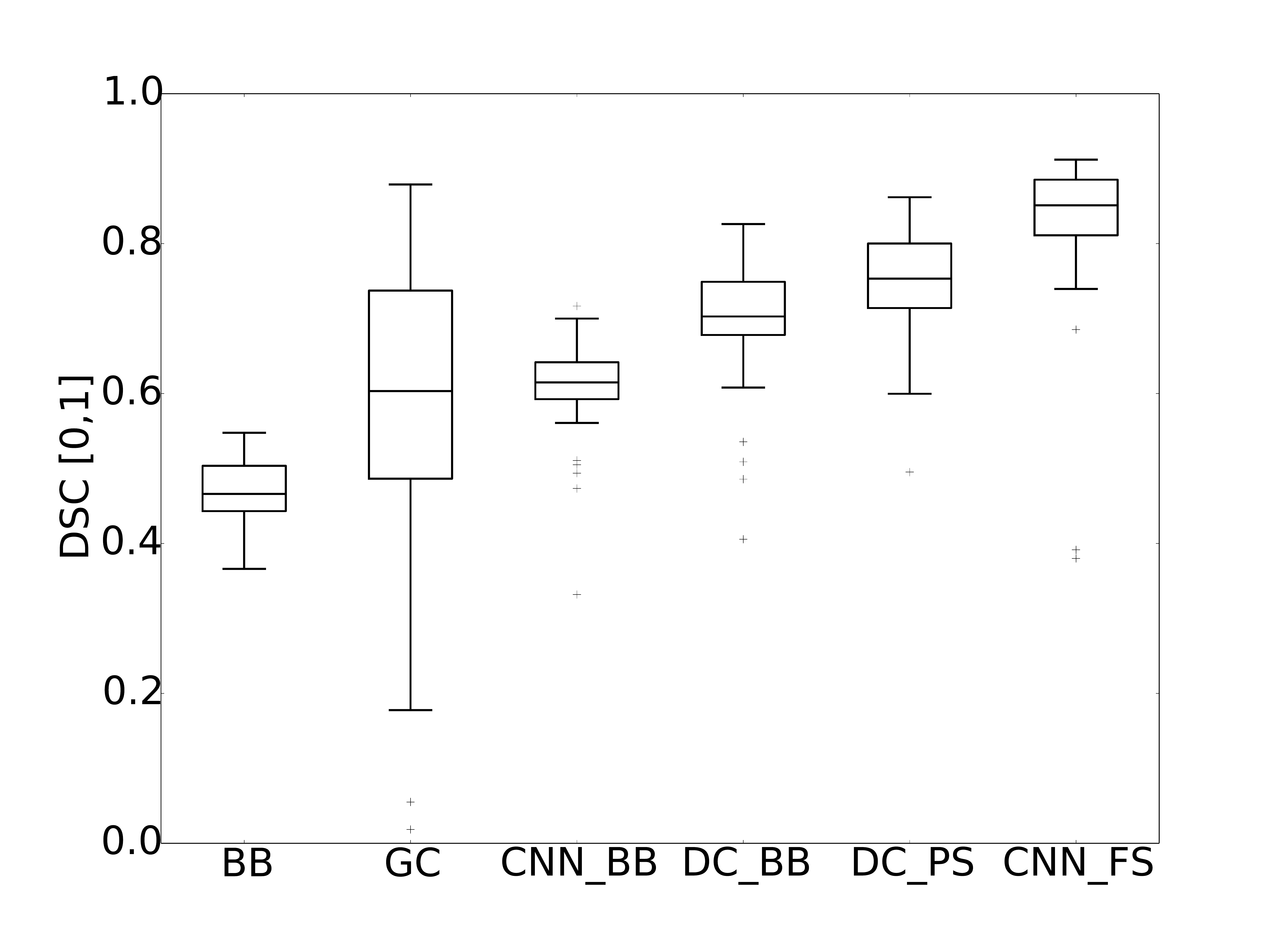} }}%
    \caption{Comparative accuracy results for the segmentation of the fetal brain (a) and lungs (b) for all methods: Initial bounding boxes (BB), GrabCut \cite{rother2004grabcut} (GC), na{\"i}ve CNN $\text{CNN}_\text{na{\"i}ve}$ learning approach from bounding boxes ($\text{CNN}_\text{BB}$), DeepCut initialised from bounding boxes ($\text{DC}_\text{BB}$), DeepCut initialised via pre-segmentation ($\text{DC}_\text{PS}$) and a fully supervised learning approach from manual segmentations ($\text{CNN}_\text{FS}$) as upper bound for this network architecture.}
    \label{fig:acc_comp}%
\end{figure*}

\begin{table}[h!]\caption{\label{ta:acc_brain} Numerical accuracy results for fetal brain segmentation. All measurements are reported as DSC [\%].}
\begin{tabular}{lccccccc}
\hline 
\hline
\noalign{\vskip 1mm} 
 & BB & GC \cite{rother2004grabcut} & $\text{CNN}_\text{na{\"i}ve}$ & $\text{DC}_\text{BB}$ & $\text{DC}_\text{PS}$ & $\text{CNN}_\text{FS}$ \\
\noalign{\vskip 1mm} 
\hline
\noalign{\vskip 1mm} 
mean & 63.0 & 80.7 & 74.0 & 86.6 & 90.3 & 94.1 \\
std. & 4.5 & 4.9 & 4.5 & 4.7 & 5.4 & 4.1 \\
\noalign{\vskip 1mm}
\hline \hline
\end{tabular}
\end{table}

\begin{table}[h!]\caption{\label{ta:acc_lungs} Numerical accuracy results for fetal lungs segmentation. All measurements are reported as DSC [\%].}
\begin{tabular}{lccccccc}
\hline 
\hline
\noalign{\vskip 1mm} 
 & BB & GC \cite{rother2004grabcut} & $\text{CNN}_\text{na{\"i}ve}$ & $\text{DC}_\text{BB}$ & $\text{DC}_\text{PS}$ & $\text{CNN}_\text{FS}$ \\
\noalign{\vskip 1mm} 
\hline
\noalign{\vskip 1mm} 
mean & 47.0 & 58.6 & 61.1 & 70.0 & 74.9 & 82.9 \\
std. & 4.1 & 19.0 & 6.4 & 8.1 & 6.7 & 10.0 \\
\noalign{\vskip 1mm}
\hline \hline
\end{tabular}
\end{table}

\section{Discussion}
The proposed \emph{DeepCut} allows for obtaining pixelwise segmentations from an image database with bounding box annotations. Its general formulation allows for readily porting it to other segmentation problems and its use of CNN models avoids feature engineering as required by other learning-based methods.

\subsection{Image Data} \label{sec:dis_imagedata}
We deliberately chose a database exhibiting large variation in the imaged anatomy (\emph{e.g.} the arbitrary position of the fetal body in the uterus, the extended gestational age range of 20-38 weeks or the presence of growth restriction (IGUR)) to test if a simple network configuration suffices for object segmentation problems constrained to bounding boxes. By restricting learning background patches from the halo $H$, we avoid learning features for the entire image domain, allowing for faster training. 

\subsection{Comparison with Related Studies}
Comparing fully supervised learning ($\text{CNN}_\text{FS}$) qualitatively, we obtain an increased mean accuracy (94.1\% DSC) over Keraudren et al. \cite{keraudren2014automated} (93.0\% DSC) and large improvements over Taleb et al. \cite{taleb2013automatic} (84.2\% DSC) for fetal brain segmentations. However, these methods \cite{keraudren2014automated,taleb2013automatic} are highly problem-specific solutions, which are applied to the entire image domain, making comparisons difficult. The only conclusion drawn from this comparison is roughly what accuracy range can be expected for automated fetal brain segmentation methods. In this sense, the generally applicable $\text{DC}_\text{PS}$ method yields similar accuracy (90.3\% DSC) by employing weak annotations, potentially placed 15x faster than pixel-wise annotations \cite{lin2014microsoft,papandreou2015weakly}.

\subsection{Differences in Brain and Lung Segmentation Performance}
For all internally compared methods, we observe a higher accuracy for brain segmentation results than for those of the lungs. There are several contributing factors to these differences, affecting all compared methods similarly. The regular shape of the brain can be better approximated with a bounding box than the lungs, which is underlined by the higher mean overlap of BB (refer to Tab. \ref{ta:acc_brain} and \ref{ta:acc_lungs}, respectively). This introduces a lower amount of false positive initial targets, facilitating training the CNN. Secondly, the contrast of the background is higher in the brain, as it is often surrounded by hyper-intense amniotic fluid or hypo-intense muscular tissue. We can observe this in the tuned CRF parameters $\theta_\beta$, penalising intensity differences (see Tab. \ref{ta:parameters}), to automatically tune to a lower value than the brain. Additionally, this can experimentally be observed with the GrabCut (GC) method, which heavily relies on intensity differences between the object and the adjacent background, performing worse than $\text{CNN}_\text{na{\"i}ve}$ (\emph{c.f.}, Fig. \ref{fig:acc_comp} (a) and (b), and Tab. \ref{ta:acc_brain} and \ref{ta:acc_lungs}).

\subsection{Effect of Initialisation on \emph{DeepCut} performance}
As shown in Fig. \ref{fig:acc_iterations}, we observe an increase in segmentation accuracy, when initialising the \emph{DeepCut} with a pre-segmentation, rather than a bounding box. Papandreou et al.  \cite{papandreou2015weakly} reported a similar increase in accuracy when initialising their EM-based algorithm. Although in both approaches, the accuracy steadily increases with the number of epochs, the methods converge to different optima. This is due to the locally optimal nature of this iterative method and other iterative optimisation schemes, such as levelsets \cite{rajchl2015variational,ukwatta20133} or iterated graphical methods \cite{nambakhsh2013left,ben2010graph,rother2004grabcut} even when employing an (approximately) globally optimal solver such as \cite{krahenbuhl2012efficient,rajchl2015variational,boykov2001fast}. Potential improvements might include to update the targets with a higher frequency than in this study (\emph{c.f.}, Tab \ref{ta:parameters}, $N_{Epochs}$ per \emph{DeepCut} iteration) or entertaining the notion, that an optimal set of CRF regularisation parameters exists at each iteration. However, tuning for the latter might be computationally expensive and thus of little practical value. Recent advances of expressing the employed CRF \cite{krahenbuhl2012efficient} as a recurrent neural network \cite{zheng2015conditional}, might be a solution for back-to-back training of the $\theta$ parameters involved in \eqref{eq:appearance_kernel} and \eqref{eq:smoothness_kernel} at each \emph{DeepCut} iteration.

\subsection{Internal Comparative Experiments}
For both lungs and brain segmentation, we report a large improvement in accuracy with \emph{DeepCut} variants over a na{\"i}ve learning approach (\emph{c.f.}, Fig. \ref{fig:acc_comp} and Tab. \ref{ta:acc_brain} and \ref{ta:acc_lungs}). As in Section \ref{sec:init}, we suggest to initialise \emph{DeepCut} with a pre-segmentation, reducing the amount of false positive targets for the initial training. A closer initialisation leads to a performance improvement, even if the pre-segmentation is not accurate (\emph{c.f.}, Fig. \ref{fig:acc_comp} (b), where there is a remarkable improvement from $\text{DC}_\text{BB}$ to $\text{DC}_\text{PS}$). The generally low standard deviations of all learning-based methods underline the robust performance compared to image segmentation methods, such as GrabCut. It can be explained, that learning a model of the object from a collection of images is favourable to fitting a model (\emph{e.g.} a GMM) to a single image, as done in many object segmentation methods. If desired, the model can be adjusted in depth to cover a wider variation of appearance and scales, as those employed in \cite{papandreou2015weakly,dai2015boxsup}. However, when the objects exhibit a large class similarity in terms of shape and appearance, simple CNN architectures could suffice for most \emph{medical} image segmentation problems.

\subsection{Conclusions}
We proposed \emph{DeepCut}, a new method to obtain pixelwise segmentations, given a database of bounding box annotations and studied variants employing an iterative dense CRF formulation and convolutional neural network models. \emph{DeepCut} is able to segment both the fetal brain and lungs from an image database of large variation in the anatomy and is readily applicable to similar problems on medical images. The proposed method performs well in terms of accuracy compared to a model trained under full supervision and simultaneously greatly reduces the annotation effort required for analysis. 

\section*{Acknowledgements}
We gratefully acknowledge the support of NVIDIA Corporation with the donation of a Tesla K40 GPU used for this research. This research was also supported by the National Institute for Health Research (NIHR) Biomedical Research Centre based at Guy's and St Thomas' NHS Foundation Trust and King's College London.  The views expressed are those of the author(s) and not necessarily those of the NHS, the NIHR or the Department of Health.  Furthermore, this work was supported by Wellcome Trust and EPSRC IEH award [102431] for the iFIND project and the Developing Human Connectome Project, which  is funded through a Synergy Grant by the European Research Council (ERC) under the European Union's Seventh Framework Programme (FP/2007-2013) / ERC Grant Agreement number 319456.

\appendices

\bibliographystyle{IEEEtran}
\bibliography{refs}

\end{document}